\documentclass[conference]{IEEEtran}
\IEEEoverridecommandlockouts
\usepackage{amsmath}
\usepackage{multirow}
\usepackage{booktabs}
\usepackage{graphicx}
\usepackage{caption}
\usepackage[numbers,sort&compress]{natbib}
\def\BibTeX{{\rm B\kern-.05em{\sc i\kern-.025em b}\kern-.08em
    T\kern-.1667em\lower.7ex\hbox{E}\kern-.125emX}}
\begin{document}

\title{The Composite Visual-Laser Navigation Method Applied in Indoor Poultry Farming Environments}

\author{
\IEEEauthorblockN{1\textsuperscript{st} Jiafan Lu}
\IEEEauthorblockA{
College of Information Science and Engineering\\
East China University of Science and Technology\\
Shanghai, China\\
Email: jiafanlu@mail.ecust.edu.cn\\}\\

\\
\IEEEauthorblockN{3\textsuperscript{rd} Yitian Ye}
\IEEEauthorblockA{
Key Laboratory of Smart Manufacturing in Energy\\
Chemical Process Ministry of Education\\
East China University of Science and Technology\\
Shanghai, China\\
Email: yitianye@mail.ecust.edu.cn\\}

\\
\IEEEauthorblockN{5\textsuperscript{th} Zixian Zhang}
\IEEEauthorblockA{
College of Information Science and Engineering\\
East China University of Science and Technology\\
Shanghai, China\\
Email: zixianzhang@mail.ecust.edu.cn\\}

\and
\IEEEauthorblockN{2\textsuperscript{nd} Dongcheng Hu$^{*}$}
\IEEEauthorblockA{
Key Laboratory of Smart Manufacturing in Energy\\
Chemical Process Ministry of Education\\
East China University of Science and Technology\\
Shanghai, China\\
Email: dongchenghu@mail.ecust.edu.cn\\}

\\
\IEEEauthorblockN{4\textsuperscript{th} Anqi Liu}
\IEEEauthorblockA{
College of Information Science and Engineering\\
East China University of Science and Technology\\
Shanghai, China\\
Email: anqiliu@mail.ecust.edu.cn\\}

\\
\IEEEauthorblockN{6\textsuperscript{th} Xin Peng$^{*}$}
\IEEEauthorblockA{
Key Laboratory of Smart Manufacturing in Energy\\
Chemical Process Ministry of Education\\
East China University of Science and Technology\\
Shanghai, China\\
Email: xinpeng@mail.ecust.edu.cn\\}
}

\maketitle

\begin{abstract}
Indoor poultry farms require inspection robots to maintain precise environmental control, which is crucial for preventing the rapid spread of disease and large-scale bird mortality. However, the complex conditions within these facilities, characterized by areas of intense illumination and water accumulation, pose significant challenges. Traditional navigation methods that rely on a single sensor often perform poorly in such environments, resulting in issues like laser drift and inaccuracies in visual navigation line extraction. To overcome these limitations, we propose a novel composite navigation method that integrates both laser and vision technologies. This approach dynamically computes a fused yaw angle based on the real-time reliability of each sensor modality, thereby eliminating the need for physical navigation lines. Experimental validation in actual poultry house environments demonstrates that our method not only resolves the inherent drawbacks of single-sensor systems, but also significantly enhances navigation precision and operational efficiency. As such, it presents a promising solution for improving the performance of inspection robots in complex indoor poultry farming settings.
\end{abstract}


\section{Introduction}
\label{sec1}

\begin{figure*}[thbp]
\centering
\includegraphics[width=2\columnwidth]{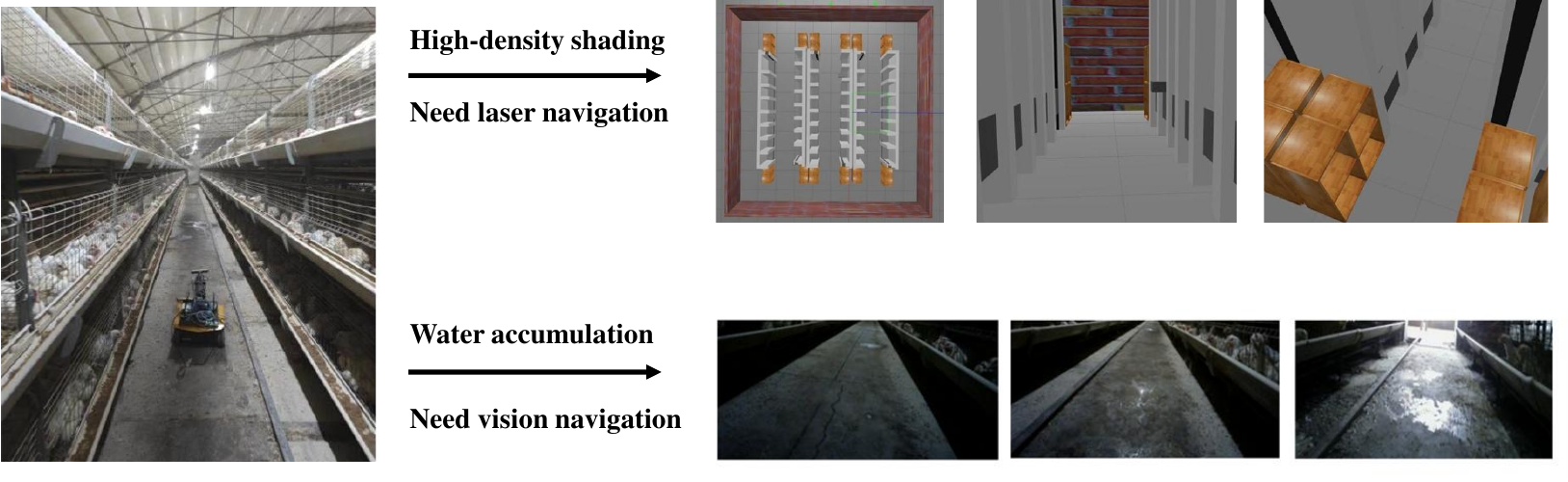}\caption{The actual environment in an indoor poultry farm, where uneven lighting and water accumulation coexist, poses significant challenges. Under these conditions, navigation based on a single sensor fails to deliver satisfactory performance.}
\label{fig_intro}
\end{figure*}

Indoor poultry farming has become the dominant approach for poultry production. In these facilities, environmental conditions, particularly temperature and humidity, are crucial for maintaining animal health\cite{aquilani2022precision}. When these conditions fall outside the optimal range, they can trigger large-scale mortality events, posing serious challenges to production efficiency and animal welfare. Traditional manual inspection methods, which rely heavily on labor-intensive processes and often yield limited accuracy, are increasingly proving insufficient to meet the growing demands of modern poultry farming\cite{ojha2015wireless}.

To address these challenges, inspection robots have been developed that can autonomously collect real-time data on temperature, humidity, and poultry health within the sheds\cite{spencer2019advances}. While these robots help to reduce the reliance on manual labor, many of them require pre-installed tracks for navigation. This reliance on fixed infrastructure not only increases the cost of retrofitting existing facilities but also leads to higher ongoing maintenance expenses, limiting their overall practicality\cite{lindqvist2020nonlinear}.

In contrast, visual navigation has emerged as a promising track-independent alternative. This method leverages either monocular or depth cameras to compute the robot’s yaw angle from captured images, thereby allowing it to adjust its pose and follow a predetermined path\cite{yang2020concrete}. Despite its simplicity, visual navigation is susceptible to variations in lighting conditions and can struggle in complex environments\cite{jiang2022thermal}. Moreover, it often necessitates the installation of visual markers or navigation lines on the ground, which can wear out over time and restrict the flexibility needed to adapt to changing layouts within the poultry house\cite{islam2023agri}.

Laser navigation represents another viable alternative\cite{chen2023sgsr}. Utilizing LiDAR sensors, this approach gathers point cloud data to accurately determine the robot's position through techniques like Adaptive Monte Carlo Localization (AMCL)\cite{pang2024low}. By integrating target point coordinates, the system calculates the yaw angle required to steer the robot toward its objective. Although laser navigation generally delivers higher accuracy compared to visual methods, it is not without its drawbacks—particularly in environments where water accumulation on the floor can lead to laser drift, thereby undermining the reliability of the navigation system.

To address these issues, we propose a visual navigation method that eliminates the need for pre-installed navigation lines, integrating it with laser navigation. This approach automatically adjusts the confidence levels of visual and laser data in response to environmental changes, ensuring accurate robot navigation even in complex settings such as poultry houses. The main contributions of this article are as follows.

\begin{enumerate}
    \item A composite visual-laser navigation method is proposed that fuses visual and laser data. Our approach harnesses the advantages of both modalities, ensuring that the robot can navigate accurately even in challenging conditions such as severe water accumulation or intense floor reflections.  
    \item A visual navigation line construction method is introduced, which employs brightness correction and edge extraction techniques. This method enables the robot to navigate without relying on physical navigation lines, thereby avoiding deviations caused by wear and tear of these lines.
    \item A fused yaw angle calculation method is developed to determine the current positional deviation. Our approach automatically evaluates the reliability of both the visual and laser-derived yaw angles and adjusts their weights accordingly. As a result, the robot is able to accurately determine its position even in complex environments.
\end{enumerate}

\section{Preliminaries}
\label{sec2}
In this section, the applications of SLAM and AMCL are reviewed, which form the foundation for the navigation approach based on fused yaw angles presented in section \ref{sec3}.

\subsection{Simultaneous Localization and Mapping}
\label{sec2A}
Simultaneous localization and mapping(SLAM) is a critical technology that enables a robot to simultaneously build a map of an unknown environment in real time and localize itself\cite{mur2015orb}. To estimate the robot's pose while constructing the map, SLAM continuously collects sensor data, extracts and matches environmental features, and employs filtering and optimization algorithms. 

In formal terms, we define the state variable $x_{1: k}=\left(x_{1}, x_{2},\ldots, x_{k}\right)$, where $x_k$ denotes the state at time $k$, and the observation variable $z_{1: k}=\left(z_{1}, z_{2},\ldots, z_{k}\right)$, where $z_k$ represents the observation at time $k$. In the visual SLAM framework used in this article, the state variable encompasses the robot’s pose, three-dimensional coordinates, and velocity at each time step, while the observation corresponds to the pixel coordinates of the feature points detected in the current frame.

The current state of the robot is estimated by maximizing the posterior probability $P$. Although it is difficult to solve for the posterior probability directly, it is feasible to solve for the optimal estimate of the state that maximizes the posterior probability at that state.
\begin{equation}
    x^{*}_\text{MAP}=\operatorname{argmax} P\left(z_{1: k}, x_{0} \mid x_{k}\right) P\left(x_{k}\right)
    \label{eq1}
\end{equation}
$x^{*}_\text{MAP}$ here represents the most likely state.

To compute the optimal state estimate, $x^{*}_\text{MAP}$, two primary approaches are typically employed: filtering methods and nonlinear optimization methods. Filtering algorithms iteratively predict the current state distribution based on the previous state and update the posterior probability using current observations. In contrast, nonlinear optimization methods jointly optimize states over multiple time steps, often requiring several iterations per optimization cycle. Given that filtering algorithms are generally computationally faster, we adopt a filtering-based approach for mapping to enhance real-time performance. 

\subsection{Adaptive Monte Carlo Localization}
\label{sec2B}
Leveraging particle filter techniques\cite{zhang2012self}, AMCL offers an enhanced approach to Monte Carlo Localization\cite{dellaert1999monte}. In this approach, each particle represents a hypothesis of the mobile platform's 2D pose at time $t$, denoted as $X_t = (x, y, \theta)$. The update of a particle's motion state depends on the robot's current control inputs and the uncertainties inherent in the motion model, typically represented as
\begin{equation}
    x_t=f(u_t,x_{t-1})+\epsilon
    \label{eq2}
\end{equation}
where $x_t$ is the state of the particle at moment $t$, $u_t$ is the control quantity, and $f(u_t,x_{t-1})$ is the motion model used to predict the position of the particle from $x_{t-1}$ to $x_t$. $\epsilon$ denotes the uncertainty of the motion, which is usually modeled as Gaussian noise.

\begin{figure}[t]
\centering
\includegraphics[width=0.85\columnwidth]{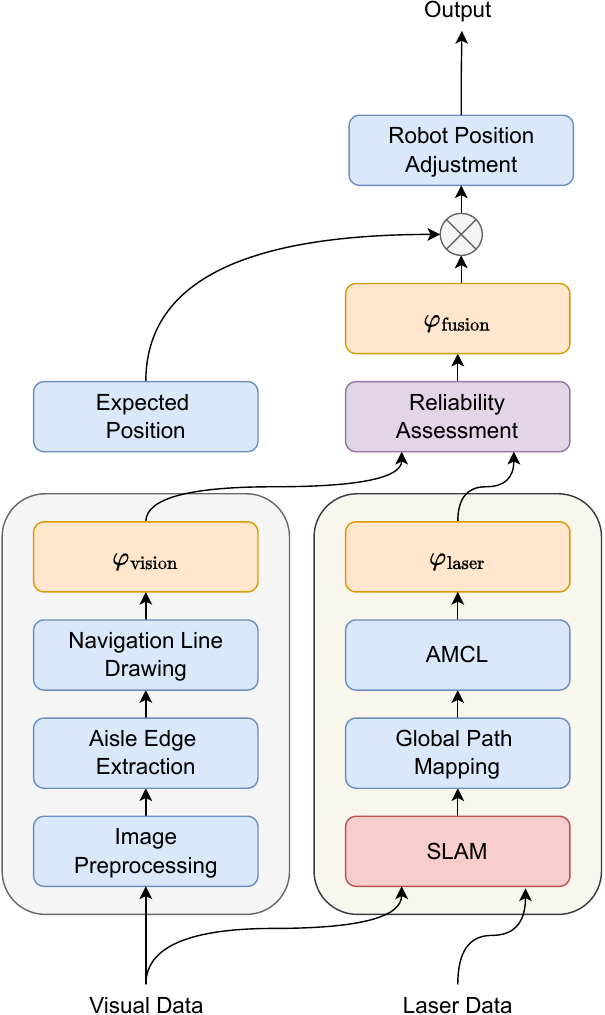}
\caption{Framework of composite vision-laser navigation.}
\label{fig_flowchart}
\end{figure}

As the robot moves, the pose of each particle is updated based on a kinematic model with control inputs. The expected observations of the predicted pose of each particle in the map are then compared with the actual observations acquired by the robot in order to update the weights of the particles according to the observation model. The particles will be resampled according to their weight distribution, and resampling is triggered when the number of effective particles drops below a specific threshold. After several iterations, the particles eventually converge around the true pose. We calculate the laser yaw angle by AMCL in section \ref{sec3}.

\section{Composite Vision-Laser Navigation}
\label{sec3}

The framework diagram of the proposed composite visual laser navigation is shown in Fig. \ref{fig_flowchart}. It can be seen that this structure consists of three main parts.

\begin{enumerate}
    \item \textit{Visual navigation line extraction:} a visual navigation line guiding the robot's target route is drawn, which is extracted by gamma correction and edge line extraction. The proposed algorithm effectively overcomes the interference of illumination.
    \item \textit{Yaw angle calculation:} visual yaw angle and laser yaw angle are calculated here, which indicate the direction of the target point.
    \item \textit{Fusion localization and navigation:} the visual yaw angle and laser yaw angle are evaluated for reliability separately, and the fusion yaw angle is obtained after weighted summation. The weights of both are automatically adjusted with the environment to keep the robot moving along the target line.
\end{enumerate}

\subsection{Visual Navigation Line Extraction}
\label{sec3A}
\begin{figure*}[htbp]
\centering
\includegraphics[width=2\columnwidth]{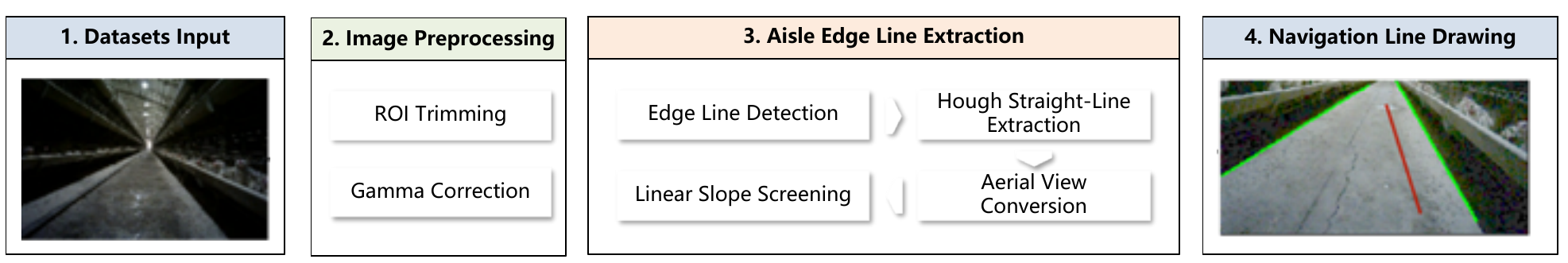}
\caption{Visual Navigation Flowchart. The actual image captured by the camera is first processed to correct brightness, then the aisle edge lines are extracted using a straight-line detection algorithm, and finally, the navigation lines are drawn.}
\label{fig_visual}
\end{figure*}

As shown in the Fig. \ref{fig_visual}, the visual navigation line extraction is divided into three steps: image preprocessing, aisle edge line extraction, and navigation line drawing.

Picture pre-processing is applied to light intensity correction to make the brightness value of the picture more uniform. This step is critical because strong sunlight near windows can cause overexposed areas when the robot is close, while images tend to be dim when it is further away. By first isolating the parts of the image that contain corridor edge lines and then applying brightness correction, the process becomes more robust.

An adaptive gamma correction method is employed to further adjust the images, making the aisle edge lines easier to extract. Gamma correction is a nonlinear transformation designed to adjust the brightness of an image to better match human visual perception. Although it is often assumed that there is a linear relationship between image brightness and display output, the actual relationship between brightness and perception is nonlinear. Thus, the core idea of gamma correction is to apply a nonlinear transformation to the image brightness, thereby optimizing both brightness and contrast. The gamma correction formula is expressed as follows:
\begin{equation}
    I_{corrected}=k\times I_{input}^{\gamma}
    \label{eq3}
\end{equation}

In this approach, $I_{corrected}$ denotes the corrected brightness value, $I_{input}$ denotes the original brightness value, and $k$ is a coefficient. The parameter $\gamma$ controls the strength of the transformation and can be computed using the following formula to achieve a more consistent brightness across different images,
\begin{equation}
    \gamma=\frac{\lg{0.5}}{\lg{(mean/255)}}
    \label{eq4}
\end{equation}
where $mean$ denotes the overall grayscale value of the image.

The extraction of aisle edge lines begins with pre-processed images. First, Canny edge detection is applied to identify potential edges, followed by the Hough transform to extract all straight lines present in the image. Given that the Hough transform is highly sensitive to the quality of edge detection, the robust noise suppression provided by Canny edge detection plays a critical role in ensuring accurate edge extraction.

In our poultry house experiments, the aisle edge lines are defined as the two lines with the maximum and minimum slopes, which correspond to the left and right boundaries of the corridor, respectively. By calculating the slopes of all detected lines, we can accurately select these key boundary lines.

Once the aisle edge lines are identified, navigation lines are drawn based on them. These navigation lines are designed to be parallel to the corridor edges, providing a clear guide for the robot's trajectory. Since the inspection task involves monitoring both sides of the corridor, two navigation lines are established for each aisle. Each navigation line is positioned at a distance equal to one-third of the total corridor width from its corresponding edge, a configuration that was chosen to balance effective coverage and navigational precision. This approach not only simplifies the navigation process but also allows for dynamic adaptation to different corridor widths, ensuring reliable operation even in complex indoor environments.

\subsection{Yaw angle calculation}
\label{sec3B}
The yaw angle refers to the angle between the robot’s orientation and the inspection trajectory. By calculating this angle, the robot can adjust its posture to ensure it remains on the predetermined inspection path and successfully completes the inspection task. In our approach, the expected straight-line inspection path is mapped onto a grid map as the global path. During operation, the robot uses a time-elastic band algorithm for local path planning and dynamically updates the local navigation targets, while the distance between the robot’s current position and the target remains constant along the global path.

\begin{figure*}[htbp]
\centering
\includegraphics[width=2\columnwidth]{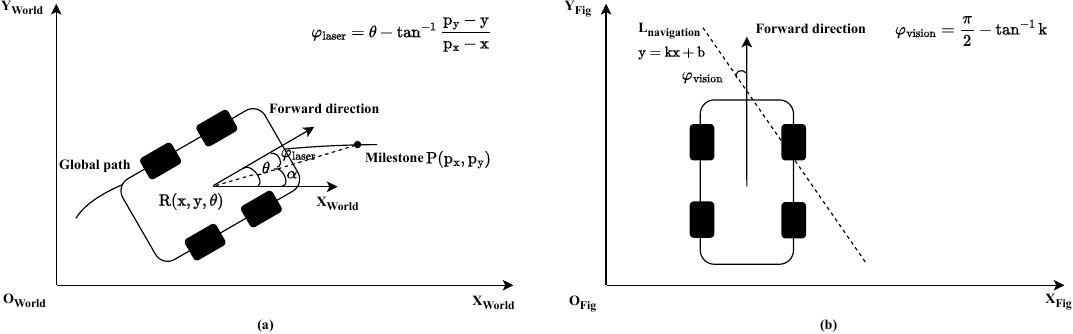}
\caption{Calculation of yaw angle. (a) Laser yaw angle, which is determined by applying a static coordinate transformation between the map and world coordinate systems; (b) Visual yaw angle, which is obtained from the visual navigation line.}
\label{fig_angle}
\end{figure*}

Fig. \ref{fig_angle} illustrates the relationship between the yaw angle and the local path in laser navigation. In an indoor poultry farming environment, the floor is relatively flat, so the robot’s pose along the z-axis is assumed to be zero. Through AMCL localization, the robot’s pose on the map is represented as $R(x,y,\theta)$, and the coordinates of the local target point during navigation are denoted as $P(p_x,p_y)$. Given the static coordinate transformation between the map and world coordinate systems, the yaw angle $\varphi_\text{laser}$ between the robot's current motion direction and the local target point can be calculated.
\begin{equation}
    \varphi_\text{laser}=\theta-\tan ^{-1} \frac{p_{y}-y}{p_{x}-x}
    \label{eq5}
\end{equation}

The visual yaw angle is computed using the navigation line. After extracting the corridor edge lines, the corridor area is segmented, and a perspective transformation is applied to convert the stereo vision image into a bird’s-eye view. The process for calculating the yaw angle is depicted in the figure, where $L_\text{navigation}$ represents the visual navigation line, $\overrightarrow{BD}$ indicates the robot’s current orientation, and $\varphi_\text{vision}$ denotes the current yaw angle. The expressions for 
$L_\text{navigation}$ and $\varphi_\text{vision}$ are given as
\begin{equation}
    L_\text{navigation}=y=kx+b
    \label{eq6}
\end{equation}
and
\begin{equation}
    \varphi_\text{vision}=\frac{\pi}{2}-\tan ^{-1} k
    \label{eq7}
\end{equation}
where $k$ is the slope of the navigation line, $b$ is the intercept, and $x$ and $y$ are the horizontal and vertical coordinates in the image coordinate system, respectively.

\subsection{Fusion localization and navigation}
\label{sec3C}
Due to the unreliability of visual navigation lines under low-light conditions and the tendency of laser measurements to drift in areas with water accumulation, we evaluate the reliability of both the visual yaw angle and the laser yaw angle separately, and then normalize them to obtain a fused yaw angle.

For the visual yaw angle, we first compute the maximum pixel value difference $d_j$ for each row
\begin{equation}
    d_{j}=i_{j, \max }-i_{j, \min } ,0 \leq j \leq m
    \label{eq8}
\end{equation}
where $m$ denotes the number of pixels along the y-axis in the image, $i_{j,\max}$ represents the rightmost white corridor pixel in the $j$th row, and $i_{j,\min}$ represents the leftmost white corridor pixel.

Next, we calculate the variance $D_V$ of $d_j$ using the average of all differences and the total number of rows $m$
\begin{equation}
    D_{V}=\sum_{j=1}^{m}\left(d_{j}-\overline{d_{j}}\right)^{2}
    \label{eq9}
\end{equation}
This variance $D_V$ is then linearly normalized to yield a reliability measure $R_\text{vision}$ that ranges between 0 and 1, with the calculation given in formula
\begin{equation}
    R_\text{vision}=\frac{D_V-D_{\min}}{D_{\max}-D_{\min}}
    \label{eq10}
\end{equation}
where $D_{\min}$ and $D_{\max}$ denote the minimum and maximum values of $d_j$, respectively.

Regarding the laser yaw angle, we assess the localization accuracy of AMCL by using an error ellipse, a region commonly employed to represent the uncertainty in estimated positions. The major and minor axes of the ellipse correspond to the standard deviations of the estimated position; therefore, a smaller error ellipse indicates higher localization accuracy. The $\mathrm{RMSE}$ is computed by 
\begin{equation}
    \mathrm{RMSE}=\sqrt{\frac{1}{n} \sum_{i=1}^{n}\left(x_{i}-y_{i}\right)^{2}}
    \label{eq11}
\end{equation}
where $x_i$ represents the estimated value at the $i$th position along the x-axis, $y_i$ represents the true value at the $i$th position along the y-axis, and localization is performed at $n$ positions. The resulting RMSE is normalized to obtain $R_\text{laser}$, with $R_{\min}$ and $R_{\max}$ representing the minimum and maximum error values and $R_t$ the current error value.
\begin{equation}
    R_\text{laser}=\frac{R_t-R_{\min}}{R_{\max}-R_{\min}}
    \label{eq12}
\end{equation}

The fused yaw angle is then derived.
\begin{equation}
   \varphi_\text{fusion}=\frac{\varphi_\text{vision} \times R_\text{vision}+\varphi_\text{laser} \times R_\text{laser}}{R_\text{vision}+R_\text{laser}}
   \label{eq13}
\end{equation}

The overall navigation process is illustrated in the Fig. \ref{fig_flowchart}. The robot receives local target points along the straight inspection path, and the fused yaw angle is used as the feedback deviation. Based on this feedback, the robot adjusts its pose accordingly, with its updated pose serving as the final output.

\section{Case Studies}
\label{sec4}
In this section, we present a detailed evaluation of the proposed method through navigation experiments conducted in a real poultry house environment. Under conditions of water accumulation and strong illumination, the composite visual-laser navigation system significantly outperforms single-mode navigation approaches.

\subsection{Experimental Setting}
\label{sec4A}
Our experiments were conducted in an actual poultry house environment using a JACKAL robot, running Ubuntu and equipped with a SICK LMS111 laser scanner and an Astra Pro depth camera. The composite visual-laser navigation system is divided into three components: visual navigation line extraction, yaw angle computation, and fused localization and navigation. We implemented visual navigation line extraction and visual yaw angle computation using the OpenCV library, while the laser yaw angle computation was developed by enhancing the ROS Navigation framework. The robot’s straight inspection path was mapped onto a grid map generated by VAGL as the global path, and its current pose was determined via AMCL to derive the laser yaw angle $\varphi_\text{laser}$.

The poultry house contains multiple rows of chicken cages, and the robot is tasked with point-based monitoring of temperature, humidity, and the condition of the poultry within the cages. Visual localization was employed to track the robot’s trajectory. An ArUco marker was affixed to the top of the robot, allowing its pose relative to a fixed camera to be calculated as it moves nearby. Since the camera’s pose relative to the ground is known, the robot’s pose relative to the ground can be further determined.

\begin{figure}[b]
\centering
\includegraphics[width=\columnwidth]{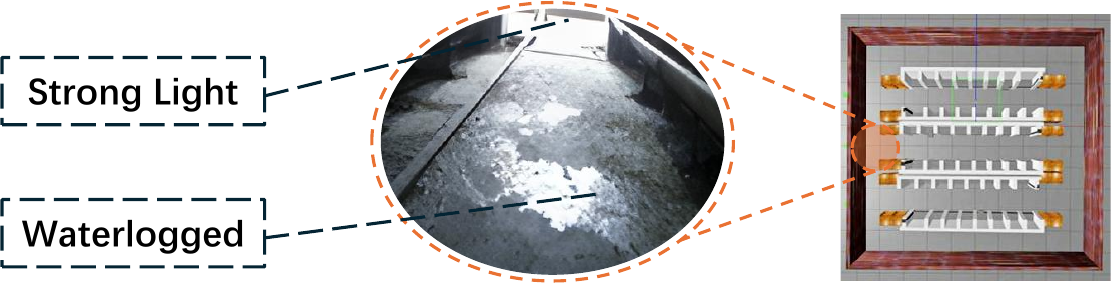}
\caption{Schematic representation of the two experimental environments in a real chicken coop. We conducted navigation experiments under bright light and in standing water, respectively.}
\label{fig_exprimental-setting}
\end{figure}

Mathematically, computing the robot’s pose relative to the camera involves determining the transformation between a point $(x,y)$ on the robot’s motion plane and a pixel $(u,v)$ in the image, which is essentially solving for the homography matrix $\mathbf{H}$ by
\begin{equation}
    \begin{bmatrix}
        x\\y\\1
    \end{bmatrix}
    =\lambda\mathbf{H}^{-1}
    \begin{bmatrix}
        u\\v\\1
    \end{bmatrix}
    \label{eq14}
\end{equation}
The homography matrix $\mathbf{H}$ consists of the camera’s intrinsic matrix $\mathbf{K}$ and the extrinsic matrix $\mathbf{T}=[\mathbf{R}\mid \boldsymbol{t}]$, where $\mathbf{R}$ is the rotation matrix (with column vectors $\boldsymbol{r_1},\boldsymbol{r_2},\boldsymbol{r_3}$) and $\boldsymbol{t}$ is the translation vector. The intrinsic matrix $\mathbf{K}$ was obtained via chessboard calibration, while the extrinsic matrix $\mathbf{T}$ was estimated using the PnP method. Since the robot operates on a horizontal plane ($z=0$), $\mathbf{H}$ can be computed using the formula:
\begin{equation}
    \mathbf{H}=\mathbf{K}
    \begin{bmatrix}
        \boldsymbol{r_1}&\boldsymbol{r_2}&\boldsymbol{t}
    \end{bmatrix}
    \label{eq15}
\end{equation}

In this manner, the robot’s real-time position is determined through the provided equation.

\subsection{Experiments under Strong Light}
\label{sec4B}
\begin{figure}[!t]
\centering
\includegraphics[width=\columnwidth]{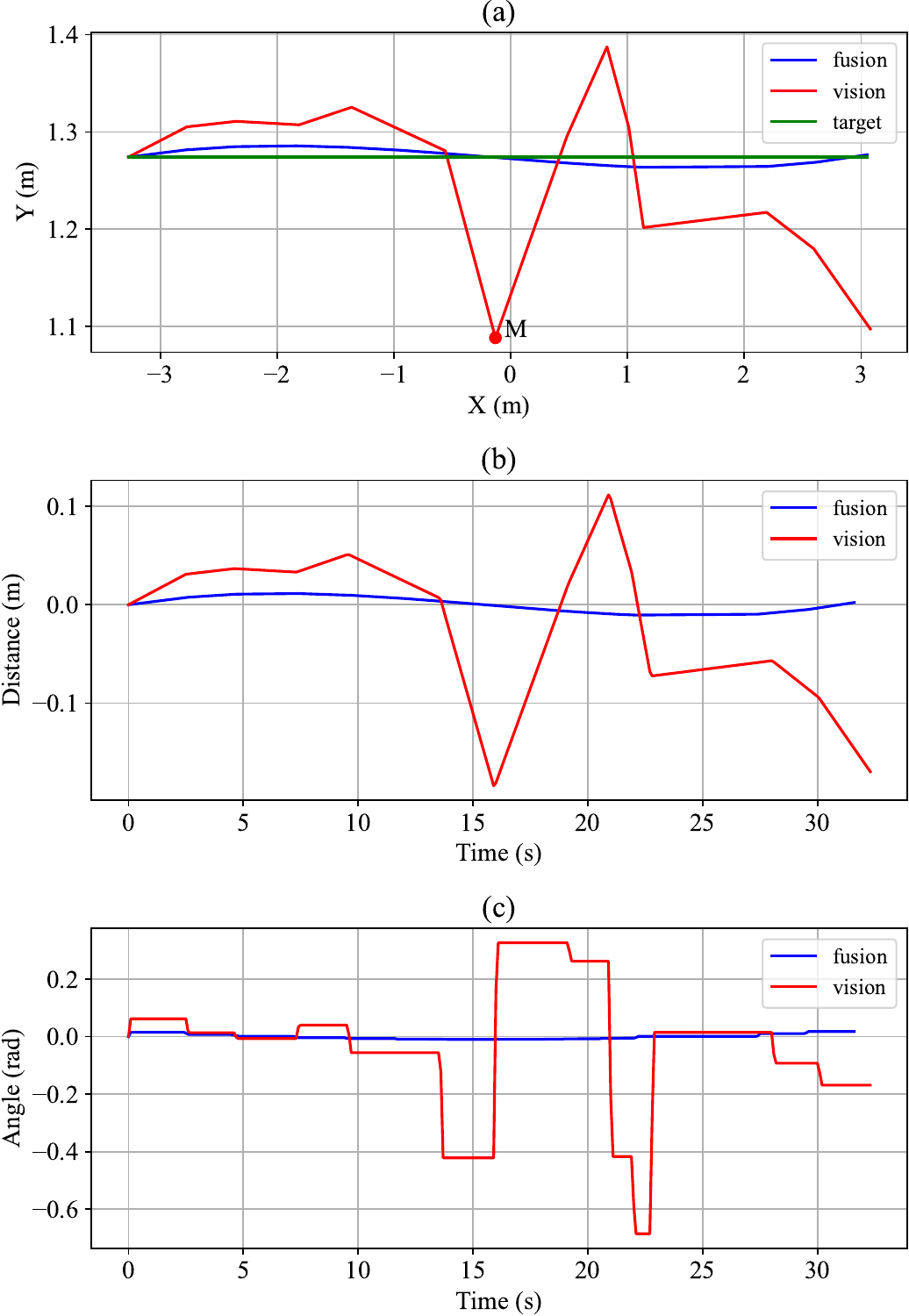}
\caption{Navigation performance of the robot in a single corridor under strong illumination. (a) Trajectory tracking plot; (b) Comparison of distance deviations; (c) Comparison of angular deviations.}
\label{fig_light-affected}
\end{figure}

The robot conducted its inspection along a path near a window, where the illumination was intense. Under these conditions, the visual navigation deviated significantly from the expected straight-line path. As shown in the Fig. \ref{fig_light-affected}(a), point M indicates the highest deviation, with the visual yaw angle $\varphi_\text{vision}$ measured at $-24.12^\circ$ and its reliability $R_\text{vision}$ at $0.05$. In contrast, the laser yaw angle $\varphi_\text{laser}$ was $1.46^\circ$ with a reliability $R_\text{laser}$ of $0.95$, demonstrating that laser navigation is less affected by strong light. The fused yaw angle $\varphi_\text{fusion}$ was $0.30^\circ$.

In strong light, the composite visual-laser navigation system increases the weight of the laser-derived yaw angle while significantly reducing the influence of the visual-derived yaw angle. As depicted in Fig. \ref{fig_light-affected}(b) and Fig. \ref{fig_light-affected}(c), the robot’s positional error was kept within $1$ cm and its angular error within $0.02$ rad, resulting in an overall trajectory that nearly coincided with the expected straight line. This confirms that the proposed algorithm effectively mitigates the impact of strong illumination.

\subsection{Experiments on Waterlogged Roads}
\label{sec4C}
\begin{figure}[!t]
\centering
\includegraphics[width=\columnwidth]{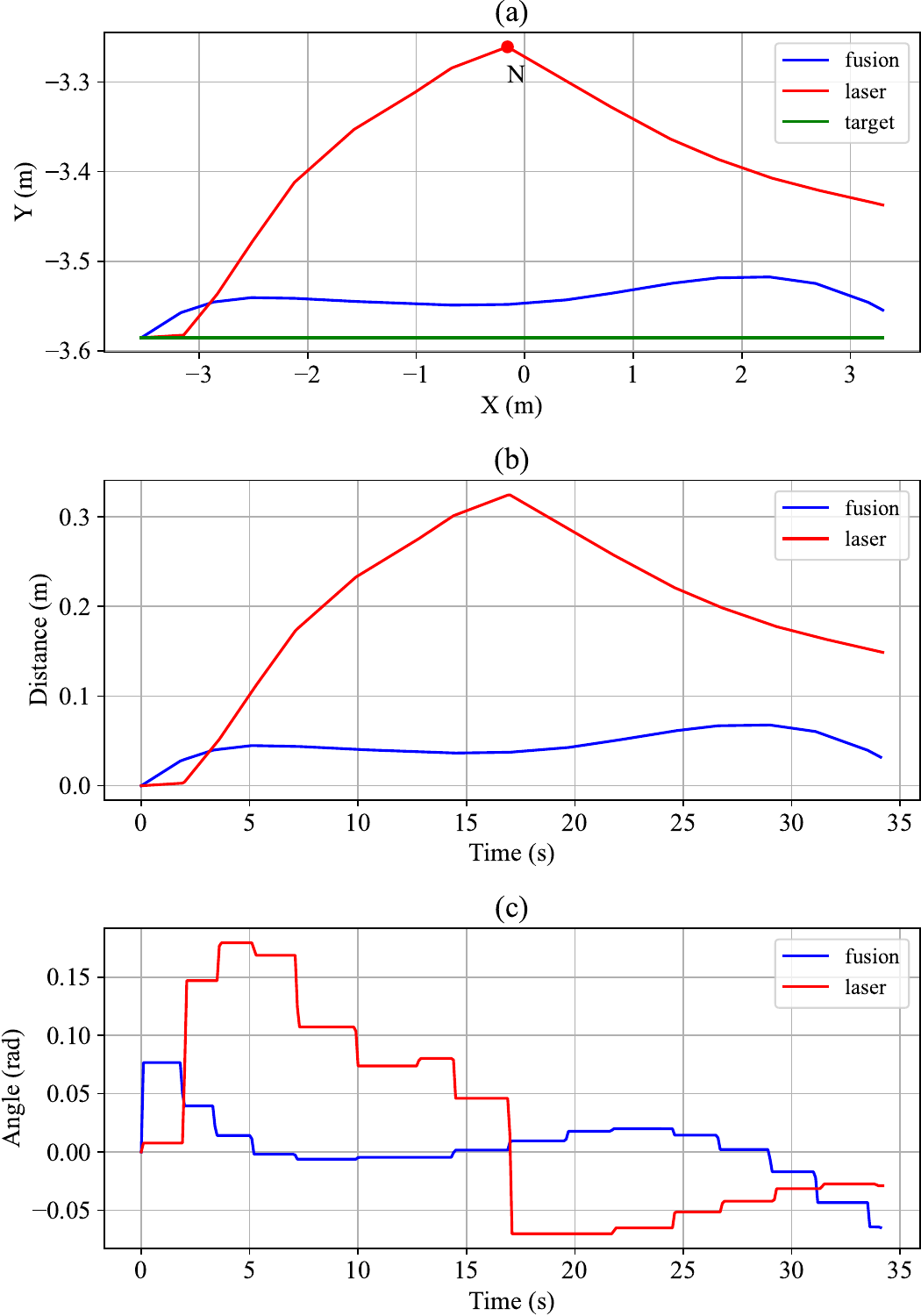}
\caption{Navigation performance of the robot in a single corridor under waterlogged conditions. (a) Trajectory tracking diagram; (b) Comparison of distance deviations; (c) Comparison of angular deviations.}
\label{fig_water-logged}
\end{figure}

When navigating a waterlogged section, the robot experienced drift in the laser measurements. As illustrated in Figure \ref{fig_water-logged}(a), point N represents the largest deviation, where the laser yaw angle $\varphi_\text{laser}$ reached $2.64^\circ$ with a reliability $R_\text{laser}$ of $0.16$, while the visual yaw angle $\varphi_\text{vision}$ was $3.50^\circ$ with a reliability $R_\text{vision}$ of $0.84$. This indicates that visual navigation is less susceptible to the effects of water accumulation. The fused yaw angle $\varphi_\text{fusion}$ was $3.36^\circ$.

\begin{table}[h]
\label{tab1}
\caption{Performance Evaluation Results in Single Aisle}
\centering
\begin{tabular}{lllll}
\toprule
Experiments                           &        & RMSE   & MAE    & MDev   \\ \midrule
\multirow{2}{*}{light-affected}       & fusion & \textbf{0.0082} & \textbf{0.0073} & \textbf{0.0115} \\
                                      & visual & 0.0859 & 0.0650 & 0.1858 \\
\midrule
\multirow{2}{*}{waterlogged}          & fusion & \textbf{0.0454} & \textbf{0.0425} & \textbf{0.0679} \\
                                      & laser  & 0.1983 & 0.1742 & 0.3250 \\ 
\bottomrule
\end{tabular}
\end{table}

In waterlogged conditions, the composite navigation system assigns a greater weight to the visual-derived yaw angle, substantially diminishing the influence of the laser-derived yaw angle. As shown in Fig. \ref{fig_water-logged}(b) and Fig. \ref{fig_water-logged}(c), the robot’s positional deviation was maintained within 7 cm and its angular deviation within $0.08$ rad, with its overall path closely matching the expected straight-line trajectory. These results demonstrate that the proposed algorithm effectively overcomes the challenges posed by water accumulation.

\section{Conclusion}
\label{sec5}
In this article, we introduce a novel composite navigation method that integrates both visual and laser data to establish a robust and reliable navigation strategy in complex environments. Unlike conventional single-sensor approaches, our method harnesses the complementary strengths of vision and laser by simultaneously collecting data from both modalities. It dynamically adjusts the reliability of each sensor based on environmental conditions and computes a fused yaw angle through a weighted summation of the individual estimates.

A key innovation of our approach lies in its adaptability: by continuously evaluating the performance of the visual and laser systems, the method ensures that the most reliable data source dominates the navigation decision, even in challenging scenarios such as strong illumination or water accumulation. Extensive experiments conducted in a real poultry house environment validate the effectiveness of the proposed method, demonstrating that it significantly outperforms traditional single-sensor navigation techniques in terms of accuracy and robustness.

These promising results underscore the potential of our composite visual-laser navigation system to enhance the operational efficiency and safety of robotic platforms in complex indoor settings, making it an attractive solution for modern agricultural and industrial applications.

\section*{Acknowledgment}
\label{Acknowledgement}
This work is supported by National Key Research and Development Program of China (2022YFB3304701), National Natural Science Foundation of China (62173145,62322303,62273214), and Shanghai AI Lab.

\bibliographystyle{ieeetr}
\bibliography{myref}

\end{document}